\documentclass[10pt,twocolumn,letterpaper]{article}

 \usepackage[pagenumbers]{cvpr} 

\usepackage[dvipsnames]{xcolor}

\definecolor{cvprblue}{rgb}{0.21,0.49,0.74}
\usepackage[pagebackref,breaklinks,colorlinks,citecolor=cvprblue]{hyperref}
\usepackage[ruled]{algorithm2e}
\usepackage{amsmath}
\usepackage{amssymb}
\usepackage{bm}
\usepackage{colortbl}
\usepackage{booktabs} 
\usepackage{color}
\usepackage{epsfig}
\usepackage{url}
\usepackage{multirow}

\usepackage{graphicx}

\usepackage{verbatim}

\usepackage{subcaption}

\usepackage{amsmath,amsfonts,bm}

\def\eqref#1{equation~\ref{#1}}

\def\1{\bm{1}}

\def\rw{{\textnormal{w}}}

\DeclareMathAlphabet{\mathsfit}{\encodingdefault}{\sfdefault}{m}{sl}
\SetMathAlphabet{\mathsfit}{bold}{\encodingdefault}{\sfdefault}{bx}{n}

\def\paperID{3047} 
\def\confName{CVPR}
\def\confYear{2024}

\title{PeLK: Parameter-efficient Large Kernel ConvNets with Peripheral Convolution}

\author{Honghao Chen$^{1,2}$\thanks{Work done during internship at Meituan Inc.}
~~
Xiangxiang Chu$^{3}$~~
Yongjian Ren$^{1,2}$~~
Xin Zhao$^{1,2}$~~
Kaiqi Huang$^{1,2,4}$\thanks{Corresponding author.}
\\[0.2cm]
$^1$CRISE, Institute of Automation, Chinese Academy of Sciences\\
$^2$School of Artificial Intelligence, University of Chinese Academy of Sciences\\
$^3$ Meituan~~
$^4$ CAS Center for Excellence in Brain Science and Intelligence Technology
}

\begin{document}
\maketitle
\begin{abstract}
Recently, some large kernel convnets strike back with appealing performance and efficiency. However, given the square complexity of convolution, scaling up kernels can bring about an enormous amount of parameters and the proliferated parameters can induce severe optimization problem. Due to these issues, current CNNs compromise to scale up to $51\times51$ in the form of stripe convolution (i.e., $51\times5+5\times51$) and start to saturate as the kernel size continues growing. In this paper, we delve into addressing these vital issues and explore whether we can continue scaling up kernels for more performance gains. Inspired by human vision, we propose a human-like peripheral convolution that efficiently reduces over 90\% parameter count of dense grid convolution through parameter sharing, and manage to scale up kernel size to extremely large. Our peripheral convolution behaves highly similar to human, reducing the complexity of convolution from $O(K^2)$ to $O(\log{K})$ without backfiring performance. Built on this, we propose Parameter-efficient Large Kernel Network (\textbf{PeLK}). Our PeLK outperforms modern vision Transformers and ConvNet architectures like Swin, ConvNeXt, RepLKNet and SLaK on various vision tasks including ImageNet classification, semantic segmentation on ADE20K and object detection on MS COCO. For the first time, we successfully scale up the kernel size of CNNs to an unprecedented $101\times101$ and demonstrate consistent improvements.
\end{abstract}    
\section{Introduction}
\label{sec:intro}

Convolutional Neural Networks (CNNs) have played a pivotal role in machine learning for decades~\cite{lecun1998gradient,alexnet,vgg, resnet}. However, their dominance has been greatly challenged by Vision Transformers (ViTs)~\cite{vit,swin,touvron2021training,pvt, chu2021twins} over recent years. Some works~\cite{raghu2021vision,vaswani2017attention} attribute the powerful performance of ViTs to their large receptive fields: Facilitated by self-attention mechanism, ViTs can capture context information from a large spatial scope and model long-range dependencies. Inspired by this, recent advances in CNNs~\cite{convnext,replk,slak} have revealed that when equipped with large kernel size (e.g., $31\times31$), pure CNN architecture can perform on par with or even better than state-of-the-art ViTs on various vision tasks.

Although large kernel convnets exhibit strong performance and appealing efficiency, a fatal problem exists: the square complexity $O(K^2)$ with respect to kernel size $K$. Due to this problem, directly scaling up kernels will bring about a huge amount of parameters. For instance, the parameter of a $31\times31$ kernel is more than 100$\times$ larger than that of a typical $3\times3$ counterpart in ResNet~\cite{resnet} and about 20$\times$ as many as that of the $7\times7$ kernel used in ConvNeXt~\cite{convnext}. The proliferated parameters subsequently induce severe optimization problem, making it useless or even harmful to directly scale up kernel size~\cite{convnext,replk,slak}. To solve, RepLKNet~\cite{replk} re-parameterize a 5$\times$5 kernel parallel to the large one to make up the optimization issue, SLaK~\cite{slak} compromise to use stripe convolution to reduce the complexity to linear and scales up to $51\times51$ (i.e., $51\times5+5\times51$). However, this is still a limited interaction range for the resolution of downstream tasks (e.g., $2048\times512$ on ADE20K) and more importantly, stripe convolution lacks the range perception of dense convolution, thus we conjecture it may undermine the model's spatial perception capacity. 

In this paper, we first conduct a comprehensive dissection of convolution forms under a unified modern framework (i.e., SLaK~\cite{slak}). We empirically verify our conjecture that dense grid convolution outperforms stripe convolution with consistent improvements across multiple kernel sizes. This phenomenon holds not only for classification task, but even more pronounced for downstream tasks, indicating the essential advantage of dense convolution over stripe form. Nevertheless, as mentioned above, the square complexity of large dense convolution leads to the proliferated \textbf{parameters}, causing rapidly increasing model size, greater optimization difficulty and thus preventing it from further scaling. This non-trivial problem naturally leads to a question: Is there a way to preserve the form of dense grid convolution while reducing the parameters required? And if so, can we further scale up dense grid convolution for more performance gains?

Unlike the dense computation of convolution or self-attention, human vision possesses a more efficient visual processing mechanism termed peripheral vision~\cite{seeing}. Specifically, human vision partitions the entire visual field into central region and peripheral region conditioned on the distance to the center of the gaze, and the number of photoreceptor cells (cones and rods) in the central region is more than 100 times that in the peripheral region~\cite{strasburger2011peripheral}. Such a physiological structure gives human vision the characteristic of blur perception: we have strong perception and see clearly in the central region, recognizing shapes and colors; whereas in the peripheral region, the visual field is blurred and the resolution decreases so we can only recognize abstract visual features such as motion and high-level contexts. This mechanism enables us to perceive important details within a small portion of the visual field ($<5\%$) while minimizing unnecessary information in the remaining portion ($>95\%$), thereby facilitating efficient visual processing in the human brain~\cite{warren1992role,rosenholtz2016capabilities,balas2009summary,deza2016can,deza2020emergent,lou2012object,rosenholtz2020demystifying,wijntjes2018context}.

Inspired by human vision and to answer the question above, we propose a novel peripheral convolution to reduce the parameter complexity of convolutions from $O(K^2)$ to $O(\log K)$ while maintaining the dense computational form. Our peripheral convolution consists of three designs: \textbf{i)} Focus and blur mechanism. We keep fine-grained parameters in the central region of the convolution kernel and use wide-range parameter sharing in the
peripheral regions; \textbf{ii)} Exponentially-increasing sharing granularity. Our sharing grid grows in an exponentially-increasing way, which is more effective than fixed granularity; \textbf{iii)} Kernel-wise positional embedding. We introduce kernel-wise positional embedding to solve the problem of detail blurring caused by wide-range peripheral sharing in an elegant and cheap way. Since our peripheral convolution dramatically reduces the parameters for large kernels (over 90\%), we are able to design large dense kernel convnets with strong performance.

Built upon the peripheral convolution above, we propose Parameter-efficient Large Kernel Network (\textbf{PeLK}), a new pure CNN architecture with Effective Receptive Field (ERF) growing exponentially with parameters. Facilitated by the elaborately designed parameter sharing mechanism, PeLK scales up kernel size at a remarkably minor parameter cost, realizing extremely large dense kernel (e.g., $51\times51$, $101\times101$) with consistent improvements. Our PeLK achieves state-of-the-art performance across a variety of vision tasks, exhibiting the potential of pure CNN architecture when equipped with extremely large kernel size.

PeLK is shown to be able to cover a much larger ERF region than prior large kernel paradigms, which we believe leads to its strong performance. More interestingly, our analysis and ablations demonstrate that the optimal design principles of peripheral convolution share striking similarities with human vision, suggesting that biologically inspired mechanisms can be promising candidates for designing strong modern networks.

\begin{figure*}[t]
  \centering
  \hspace{8pt}
    \begin{subfigure}{0.28\textwidth}
      \centering   
      \includegraphics[width=1\linewidth]{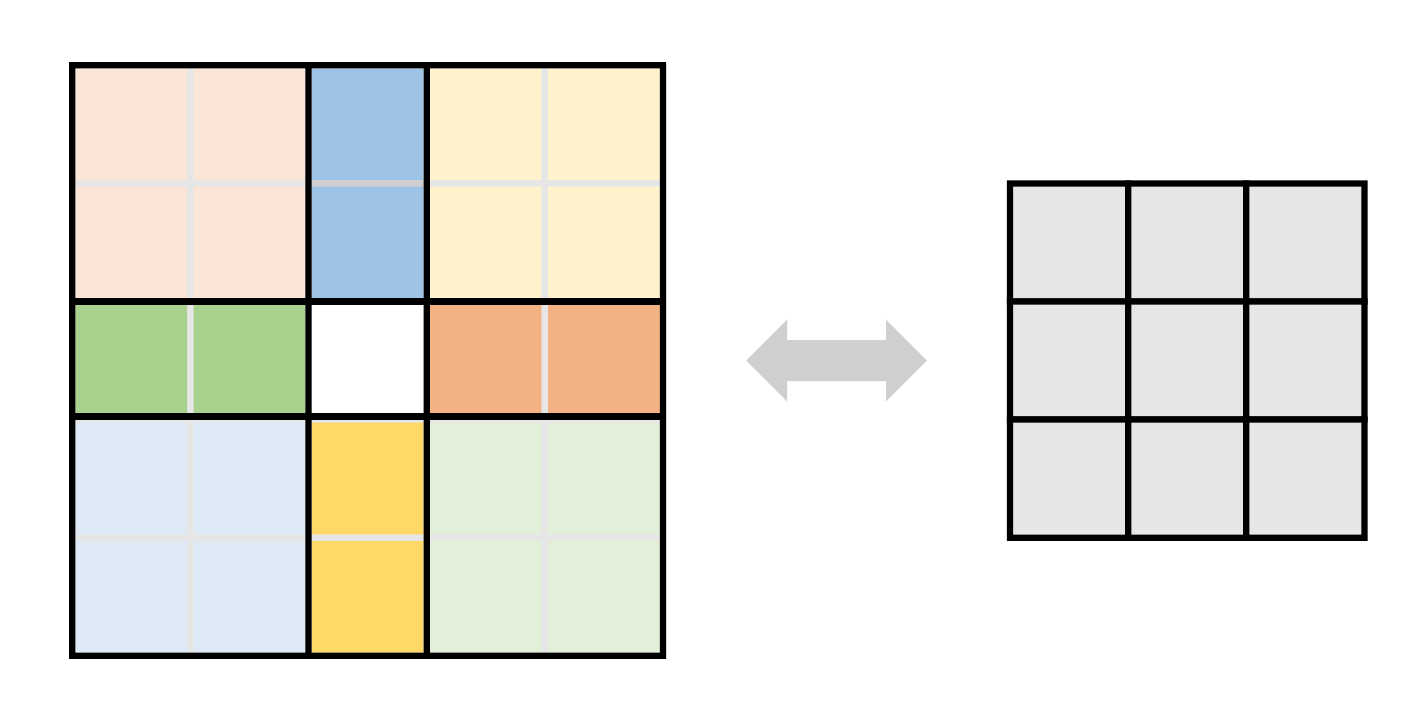}
        \caption{Parameter Sharing.}
        \label{fig:left}
    \end{subfigure}   
    \hspace{29pt}
    \begin{subfigure}{0.6\textwidth}
      \centering   
      \includegraphics[width=1\linewidth]{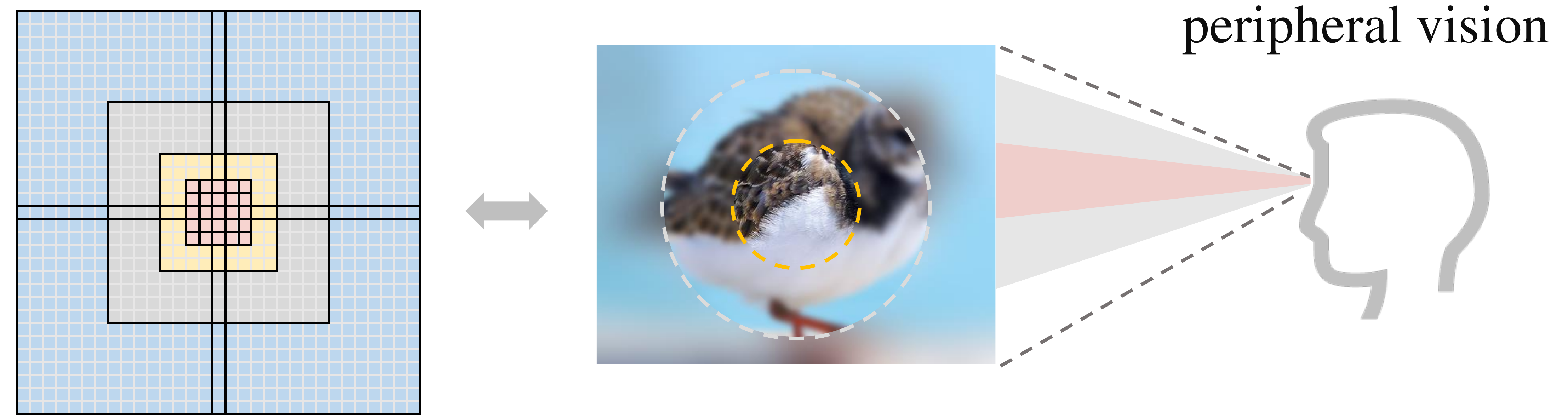}
        \caption{Peripheral Convolution.}
        \label{fig:right}
    \end{subfigure}
\caption{
\label{fig:total}
\textbf{(a) Illustration of parameter sharing.} Using a 3$\times$3 convolution to parameterize a 5$\times$5 convolution, the positions with the same color share the same parameter. The corresponding sharing grid is $[2,1,2]$. \textbf{(b) Illustration of peripheral convolution.} Our sharing grid contains two designs: i) focus and blur mechanism; ii) exponentially-increasing sharing grid.
}
\end{figure*}

\section{Related Work}
\label{sec:related}

\subsection{Large Kernel Convolutional Networks} 
Large kernel convolutional networks can date back to a few old fashion models from the early days of deep learning~\cite{alexnet,szegedy2015going,szegedy2016rethinking}.  After VGG-Net~\cite{vgg}, it becomes a common practice to use a stack of small kernels (e.g., $1\times1$ or $3\times3$) to obtain a large receptive field over the past decade. Global Convolutional Network (GCNs)~\cite{gcn} enlarges the kernel size to 15 by employing a combination of stripe convolutions ($1\times {\rm M}$ + ${\rm M}\times1$) to improve the semantic segmentation task. However, the proposed method is reported to harm the performance on ImageNet. Recently, large kernel convnets strike back with appealing performance~\cite{convmixer,convnext,replk,slak}. ConvMixer~\cite{convmixer} use $9\times9$ depthwise convolution to replace the spatial mixer of ViT~\cite{vit} and MLP-Mixer~\cite{mlpmixer} (i.e., self-attention block and fully-connection block respectively). ConvNeXt~\cite{convnext} aligns with Swin's~\cite{swin} design philosophy to explore a strong modern CNN architecture equipped with $7\times7$ depthwise convolution. RepLKNet~\cite{replk} impressively scales up the kernel size to $31\times31$ by re-parameterizing a small kernel (e.g., $5\times5$) parallel to it and performs on par with Swin Transformer~\cite{swin}. Our work is also inspired by LargeKernel3D~\cite{largekernel3d}, which introduces large kernel design into 3D networks and scales up to $17\times17\times17$. In contrast, we explore the extremety of 2D universal convolution, scaling up to a much larger $101\times101$ in a human-like pattern.  SLaK~\cite{slak} combines decomposed convolution with dynamic sparsity to scale up kernels to $51\times51$ in the form of stripe convolution (e.g., $51\times5$ + $5\times51$). However, it starts to saturate as the kernel size continuous growing. Different from those prior arts, we investigate which kind of convolution form is more effective in large kernel designs. More importantly, we explore the design of extremely large dense kernel and test whether it can bring further gains.
\subsection{Peripheral Vision for Machine Learning}
Human vision has a special visual processing system termed peripheral vision~\cite{seeing}. It partitions the entire visual field into multiple contour regions depending on the distances to the fovea, each characterized by a distinct resolution granularity for recognition. The work of Rosenholtz~\cite{rosenholtz2016capabilities} discusses in depth important findings and existing myths about peripheral vision, suggesting that peripheral vision is more crucial to human perception on a range of different tasks than previously thought. Following this, many studies~\cite{balas2009summary,deza2016can,deza2020emergent,lou2012object,rosenholtz2020demystifying,wijntjes2018context} have been devoted to uncovering the underlying principles and deep implications of peripheral vision mechanisms. Since peripheral vision plays such a vital role in human vision, a number of pioneering works~\cite{deza2020emergent,fridman2017sideeye,gould2007peripheral,harrington2021finding,lukanov2021biologically,wang2017central} dig into  the linkage between peripheral vision and machine vision (e.g., CNNs). ~\cite{vuyyuru2020biologically} introduces a biologically-inspired mechanism to improve the robustness of neural networks to small adversarial perturbations. FoveaTer~\cite{foveater} uses radial-polar pooling regions to dynamically allocate more fixation/computational resources to more challenging images. PerViT~\cite{pervit} proposes to incorporate peripheral position encoding to the multi-head self-attention layers to partition the visual field into diverse peripheral regions, showing that the network learns to perceive visual data similarly to the way that human vision does. Continuing previous study, this paper explores to blending human peripheral vision with large kernel convnets, and introduces a novel peripheral convolution to efficiently reduce dense convolution's parameters.
\section{Dense Outperforms Stripe Consistently}\label{sec3}

We first investigate whether dense grid convolutions are better than stripe convolutions. We take a unified modern framework SLaK~\cite{slak} to conduct this study. According to RepLKNet~\cite{replk}, large kernel convolution boosts downstream tasks much more than ImageNet classification. So we not only evaluate on ImageNet-1K but also on ADE20K as our benchmark. We adopt the efficient large-kernel implementation developed by MegEngine~\cite{MegEngine} in this paper.

Following SLaK~\cite{slak}, we train all models for a 120-epoch schedule on ImageNet. The data augmentations, regularization and hyper-parameters are all set the same. We then use the pretrained models as the backbones on ADE20K. Specifically, we use the UperNet~\cite{xiao2018unified} implemented by MMSegmentation~\cite{mmseg2020} with the 80K-iteration training schedule. We do not use any advanced techniques nor custom algorithms since we seek to evaluate the backbone only.

SLaK introduce a two-step recipe for scaling up kernel to $51\times51$: 1) Decomposing a large kernel into two rectangular, parallel kernels; 2) Using dynamic sparsity and expanding more width. In order to thoroughly analyze the effect of convolution form, we conduct experiments both w/ and w/o sparsity. By default, we re-parameterize a $5\times5$ convolution to ease the optimization problem as taken by SLaK and RepLKNet. The results of Table~\ref{table-dense1} show that dense grid convolution exceeds stripe convolution regardless of dynamic sparsity.

\begin{table}[b]
\vspace{-1.5mm}
	\begin{center}
  \caption{Comparison w/ and w/o dynamic sparsity. Dense convolution outperforms stripe convolution both on ImageNet and ADE20K.} 
  \vspace{-2.0mm}
   \label{table-dense1}
		\small
		\begin{tabular}{lcccl}
			\toprule
			Method			&	Kernel	&	Spasity	&	Acc & mIoU\\
			\midrule
			SLaK-51       &   51$\times$5 + 5$\times$51       &   w/          &81.6  & 46.5\\
			RepLK-51       &   51$\times$51    &   w/          &81.7 &46.9 (+0.4)\\
			\midrule
			SLaK-51       &   51$\times$5 + 5$\times$51       &   w/o          &81.3  & 46.1\\
			RepLK-51       &   51$\times$51    &   w/o          &81.6 &46.6 (+0.5)\\
			\bottomrule
		\end{tabular}
	\end{center}
\vspace{-6.5mm}
\end{table}
 
\begin{figure}[t]
\centering
\includegraphics[width=\linewidth]{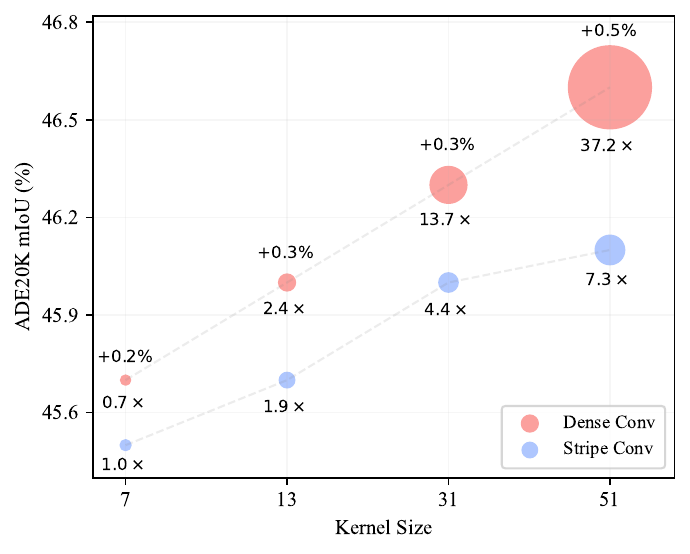} 
\vspace{-6.5mm} 
\caption{\textbf{Comparison under different kernel sizes.} We depict the mIoU gains on ADE20K and the multiple of convolutional parameters. Dense grid convolution exceeds stripe convolution consistently but brings rapidly-increasing parameters.}
 \label{dense_vs_stripe}
\vspace{-3.9mm} 
\end{figure}

We further explore convolution forms (i.e., K$\times$K v.s. K$\times$N) under different kernel sizes. Specifically, we fix the shorter edge of SLaK's stripe conv to be 5 as the default setting (N=5), and then gradually decrease K from 51 to 7. We do not use dynamic sparsity to give a sheer ablation on convolutional forms. As shown in Fig.~\ref{dense_vs_stripe}, dense grid convolution outperforms stripe convolution consistently among multiple kernel sizes and the gains increase with the kernel size, demonstrating the essential advantage of dense grid large kernel convolution.

Nevertheless, as discussed in Section~\ref{sec:intro}, the square complexity of dense grid convolution can bring about proliferated parameters. For instance, as shown in Fig.~\ref{dense_vs_stripe}, scaling up kernel from 7 to 51 only bring about 7.3$\times$ params for stripe conv while that for dense conv is 53.1$\times$. Given that the human's peripheral vision has only a minimal number of photoreceptor cells in the peripheral regions, we argue that dense parameters are not necessary for peripheral interactions. Motivated by this, we seek to reduce parameter complexity by introducing the peripheral vision mechanism while preserving the dense computation to keep dense convolution's strong performance.

\section{Parameter-efficient Large Kernel Network}

\subsection{Peripheral Convolution}
Formally, a standard 2D convolution kernel consists of a 4-D vector: $\rw \in \mathbb{R}^{c_{\rm in}\times c_{\rm out}\times \rm k\times \rm k}$, where $c_{\rm in}$ stands for input channels, $c_{\rm out}$ is output channels, and $\rm k$ means the spatial kernel dimension. We seek to parameterize $\rw$ by a smaller kernel $\rw_{\theta} \in \mathbb{R}^{c_{\rm in}\times c_{\rm out}\times \rm k'\times \rm k'}$ through spatial-wise parameter sharing, where $0<\rm k'\leq \rm k$. 


Firstly, we define the sharing grid $S=[s_0, s_1,...,s_{\rm k'-1}]$, where $\sum_{i=0}^{\rm k'-1}s_i={\rm k}$. According to $S$, we partition the $\rm k\times \rm k$ positions into $\rm k'\times \rm k'$ regions:

for $a, b=0, 1, ..., \rm k'-1,$
\begin{equation}\label{equation1}
Z_{a,b} = \left\{ (x,y) \Bigg| \sum_{i=0}^{a-1}s_i \leq x<\sum_{i=0}^{a}s_i, \sum_{j=0}^{b-1}s_j \leq y<\sum_{j=0}^{b}s_j  \right\}
\end{equation}

For brevity, we stipulate that $\sum_{i=0}^{-1}s_i=0$ in Eq.~\ref{equation1}. Then for any position $(x,y)\in Z_{a,b}$, we set $\rw(x,y)=\rw_{\theta}(a,b)$. In this way, we can utilize a small kernel to parameterize a much larger kernel, achieving spatial-wise parameter sharing. Fig.~\ref{fig:left} depicts the illustration of this design. 

Next, we elaborate on the key designs of our peripheral convolution. We denote the kernel radius of $w_{\theta}$ as $r$. For easier comprehension, here we reformulate the sharing grid into an axisymmetric form: $S=[\bar{s}_{-r},\bar{s}_{-r+1},...,\bar{s}_{-1},\bar{s}_0,\bar{s}_1,...,\bar{s}_{r-1},\bar{s}_r]$, where $r=\frac{\rm k'-1}{2}$.

Akin to human's peripheral vision, the sharing grid of our peripheral convolution mainly consists of two core designs: \textbf{i) Focus and blur mechanism.} As shown in Fig.~\ref{fig:right}, We keep fine-grained parameters in the central region of the convolution kernel, where the sharing grid is set to 1 (i.e., not sharing). For the peripheral region, we utilize large-range parameter sharing to exploit the spatial redundancy of peripheral vision. We demonstrate in Section~\ref{ablation} that the fine granularity in the central region is of vital importance, while the peripheral region can withstand a wide range of parameter sharing without backfiring performance; \textbf{ii) Exponentially-increasing sharing granularity.} Human vision declines in a quasi-exponential mode~\cite{pramod2022human}. Inspired by this, we design our sharing grid to grow in an exponentially-increasing way. This design can elegantly reduce the parameter complexity of convolution from $O(K^2)$ to $O(\log{K})$, making it possible to further enlarge dense convolution's kernel size. Specifically, the sharing grid $S$ is constructed by:
\begin{equation}
\bar{s}_i=\left\{
             \begin{array}{ll}
             1, & \quad{\rm if} \ \ |i|\leq r_c \\
             {\rm m}^{(|i|-r_c)}, & \quad{\rm if} \ \ r_c<|i|\leq r  
             \end{array}
\right.
\end{equation}
where $r_c$ is the radius of the central fine-grained region, $\rm m$ is the base of the exponential growth
and $\rm m$ is set to 2 by default.
\begin{figure}[t]
\centering
\includegraphics[width=\linewidth]{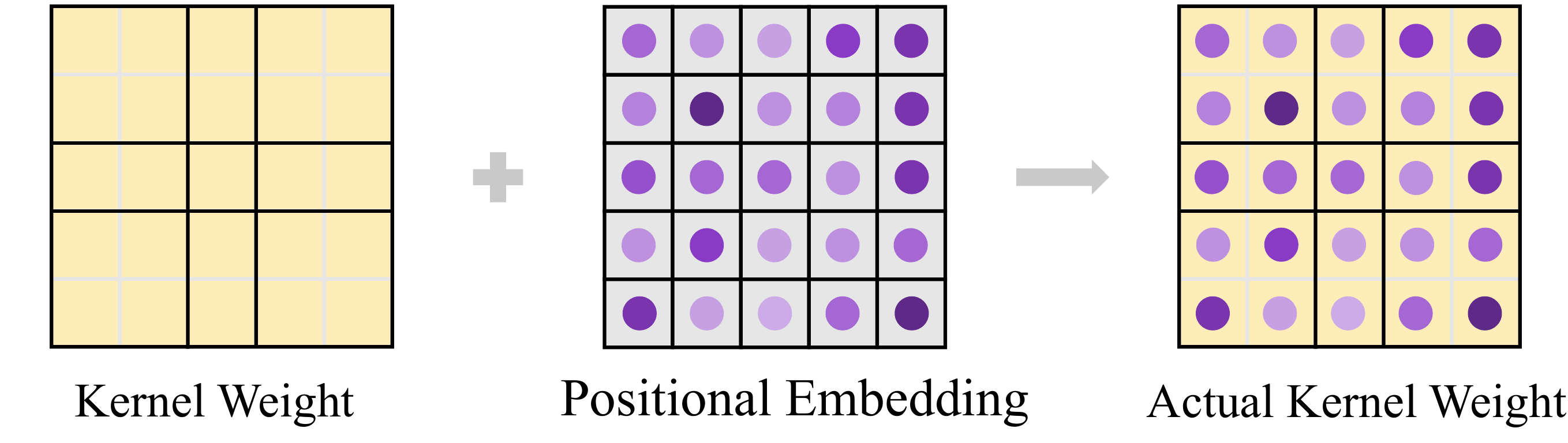} 
\caption{\textbf{Illustration of kernel-wise positional embedding.} The position embedding enables the kernel to distinguish specific positions in the sharing region, making up the detail-capturing ability of large kernels.}
 \label{fig2}
\vspace{-1.9mm} 
\end{figure}

\subsection{Kernel-wise Positional Embedding}
Despite that the proposed peripheral convolution effectively reduces the parameters for dense convolution, the large range of parameter sharing may bring another issue: local detail blurring in peripheral regions. Especially when the kernel size is scaled up to more than 50 or even 100 in the form of peripheral convolution, this phenomenon will be further amplified when a single parameter needs to process $8\times8$ or even $16\times16$ peripheral regions.

To solve, we propose the kernel-wise positional embedding. Formally, given a set of input features $X$, We process these features by a convolution with kernel weights ${\rm w} \in\mathbb{R}^{c_{\rm in}\times c_{\rm out}\times \rm k\times \rm k}$. We initialize the position embedding ${\rm h}\in\mathbb{R}^{c_{\rm in}\times \rm k\times \rm k}$ with \textit{trunc normal}~\cite{timm} initialization. The convolution process at the output position $(x,y)$ can be represented as:

\begin{equation}
Y(x,y) = \sum_{i=-r_{\rm w}}^{r_{\rm w}}\sum_{j=-r_{\rm w}}^{r_{\rm w}} \rw(i,j)\cdot\big( X(x+i,y+j)+ {\rm h}(i,j)\big)
\end{equation}

where $Y$ is the output. $\rm r_{\rm w}$ is the radius of the kernel $\rm w$ and we have $r_{\rm w}=\frac{\rm k-1}{2}$.

As illustrated in Fig.~\ref{fig2}, by introducing kernel-wise positional embedding for kernel, we can distinguish specific locations in shared areas, so as to make up for the problem of vague local details caused by sharing. Actually, this can be viewed as adding bias with relative position information to the input features. It is worth noting that all the kernels in a stage share the same positional embedding $\rm h$, thus the additional parameters brought by $\rm h$ are negligible. This design solves the position insensitivity problem caused by sharing weights in a cheap and elegant way, especially for extremely large kernels, e.g., $51\times51$ and $101\times101$.

\subsection{Partial Peripheral Convolution}
Large kernel convnets have been shown to have high channel redundancy~\cite{inceptionnext} and suit well with sparsity~\cite{slak}. Since our peripheral convolution enables us to design larger dense convolution with stronger spatial perception ability, we hope to further exploit the channel redundancy of large convolution. We introduce an Inception-style design where only partial channels of the feature map will be processed by convolution. We follow a simple philosophy: more identity mapping to exploit the channel redundancy. Specifically, for input $X$, we split it into two groups along the channel dimension,
\begin{equation}
\begin{aligned}
X_{\text{conv}}, X_{\text{id}} &= \text{Split} (X)\\
&=X_{:,:,:g}, X_{:,:,g:}
\end{aligned}
\end{equation}
where g is the channel numbers of convolution branches and set to $\frac{3}{8}C_{in}$ by default. Then the split inputs are fed into peripheral convolution and identity mapping respectively,
\begin{equation}
\begin{aligned}
    X_{\text{conv}}^{'} &= \text{Peripheral Conv}(X_{\text{conv}}) \\
    X_{\text{id}}^{'} &= X_{\text{id}}
    \end{aligned}
\end{equation}
Finally, the outputs from two branches are concatenated to restore the original shape,
\begin{equation}
    X^{'} = \text{Concat}(X_{\text{conv}}^{'}, X_{\text{id}}^{'}).
\end{equation}

This design can be seen as a special case of Inception-style structure, such as Inception~\cite{inception}, Shufflenet~\cite{shufflenet, shufflenetv2} and InceptionNeXt~\cite{inceptionnext}. They utilize different operators in parallel branches while we take a much simpler philosophy: only peripheral convolution and identity mapping. We empirically find that this design suits well for peripheral convolutions with extremely large kernels, significantly reducing FLOPs without backfiring performance.

\subsection{Architecture Specification}
Built on the above designs and observations, we now elaborate the architectures of our Parameter-efficient Large Kernel Network (PeLK). We mainly follow ConvNeXt and SLaK to construct models with several sizes. Specifically, PeLK also adopts a 4-stage framework. We build the stem with a convolution layer with $4\times4$ kernels and 4 stride. The block numbers of stages are $[3, 3, 9, 3]$ for tiny size and $[3, 3, 27, 3]$ for small/base size. The kernel sizes for PeLK's different stages are $[51,49,47,13]$ by default. For PeLK-101, the kernel sizes are scaled up to $[101,69,67,13]$.

By default, we keep the central $5\times5$ region to be fine-grained. For PeLK-101, we enlarge the central region to $7\times7$ to adjust the increased kernel. Following SLaK, we also use dynamic sparsity to enhance model capacity. All the hyperparameters are set the same ($1.3\times$ width, 40\% sparsity). We give thorough ablations for kernel configurations in section~\ref{ablation}. 

\section{Experiments}
 In this section, we first conduct experiments on various essential vision tasks to evaluate PeLK with state-of-the-art baselines. Then in section~\ref{ablation} we comprehensively ablate on the design principles of our peripheral convolution.

\subsection{Semantic Segmentation}
\label{4p2}
For semantic segmentation, we evaluate PeLK backbones on the ADE20K benchmark~\cite{ade20k}, which consists of 25K images and 150 semantic categories. We use the UperNet~\cite{upernet} task layer for semantic segmentation. Following Swin and ConvNeXt, We train Upernet for 160K iterations with single-scale inference. The results are reported in Table~\ref{semantic segmentation} with mean Intersection of Union (mIoU) as the evaluation metric. Our proposed PeLK exceeds previous state-of-the-art models with remarkable improvements, demonstrating the effectiveness of our framework.
\begin{table}[t]
	\centering
  \caption { Semantic segmentation comparison on ADE20K of different methods. We report the single-scale mIoU following ConvNeXt and SLaK. FLOPs are based on input sizes of (2048, 512).}
 \label{semantic segmentation}
	\scalebox{0.87}{
	\begin{tabular}[c]{lcccc}
		\toprule
            {\multirow{2}{*}{Method}} & Kernel & Params & FLOPs & mIoU\\  
            {} & size  & (M) & (G) & (\%)\\
		\midrule
  Swin-T~\citep{swin}& N/A & 60 & 945 & 44.5\\
  ConvNeXt-T~\citep{convnext}& 7-7-7-7 & 60 & 939 & 46.0\\
  SLaK-T~\citep{slak}& 51-49-47-13 & 64 & 957 & 47.6\\
  \rowcolor[rgb]{0.95,0.95,0.95} \textbf{PeLK-T} & 51-49-47-13 & 62 & 970 &  \textbf{48.1}\\
  \midrule
  Swin-S~\citep{swin}& N/A & 81 & 1038 & 47.6\\
  ConvNeXt-S~\citep{convnext}& 7-7-7-7 & 82 & 1027 & 48.7\\
  SLaK-S~\citep{slak}& 51-49-47-13 & 89 & 1057 & 49.4\\
  \rowcolor[rgb]{0.95,0.95,0.95} \textbf{PeLK-S} & 51-49-47-13 & 84 & 1077 &  \textbf{49.7}\\
  \midrule
  Swin-B~\citep{swin}& N/A & 121 & 1188 & 48.1\\
  ConvNeXt-B~\citep{convnext}& 7-7-7-7 & 122 & 1170 & 49.1\\
  RepLKNet-B~\citep{replk}& 31-29-27-13 & 112 & 1170 & 49.9\\
  SLaK-B~\citep{slak}& 51-49-47-13 & 131 & 1210 & 50.2\\
  \rowcolor[rgb]{0.95,0.95,0.95} \textbf{PeLK-B} & 51-49-47-13 & 126 & 1237 &  \textbf{50.4}\\
   \rowcolor[rgb]{0.95,0.95,0.95} \textbf{PeLK-B-101} & 101-69-67-13 & 126 & 1339 &  \textbf{50.6}\\
		\bottomrule
	\end{tabular}
	}
\end{table}
\subsection{Object Detection}
For object detection/segmentation, we conduct experiments with Cascade Mask R-CNN~\cite{mask-r-cnn,cascade} on MS-COCO~\cite{coco}. Following ConvNeXt, we use the multi-scale setting and default configurations in MMDetection~\cite{mmdet}. The Cascade Mask R-CNN model is trained with the 3x (36-epoch) training schedule. As shown in Table~\ref{object detection}, PeLK achieves higher mAP than state-of-the-art methods, samely validating our superiority.

\begin{table}[t]
	\centering
  \caption {Object detection comparison on COCO of different methods. FLOPs are based on input sizes of (1280, 800).}
 \label{object detection}
	\scalebox{0.92}{
	\begin{tabular}[c]{lcccc}
		\toprule
            {\multirow{2}{*}{Method}} & Params & FLOPs & {\multirow{2}{*}{$\rm{AP^ \text{box} }$}} & {\multirow{2}{*}{$\rm{AP^\text{mask} }$}}\\  
            {} & (M)  & (G) & {} & {}\\
		\midrule
  Swin-T~\citep{swin}& 86 & 745 & 50.5 & 43.7\\
  ConvNeXt-T~\citep{convnext}& 86 & 741 & 50.4 & 43.7\\
  \rowcolor[rgb]{0.95,0.95,0.95} \textbf{PeLK-T} & 86 & 770 & \textbf{51.4} &  \textbf{44.6}\\
  \midrule
  Swin-S~\citep{swin}& 107 & 838 & 51.8 & 44.7\\
  ConvNeXt-S~\citep{convnext}& 108 & 827 & 51.9 & 45.0\\
  \rowcolor[rgb]{0.95,0.95,0.95} \textbf{PeLK-S} &  108 & 874 & \textbf{52.2} &  \textbf{45.3}\\
  \midrule
  Swin-B~\citep{swin}& 145 & 982 & 51.9 & 45.0\\
  RepLKNet-B~\citep{replk}& 137 & 965 & 52.2 & 45.2 \\
  SLaK-B~\citep{slak}& 152 & 1001 & 52.5 & 45.5\\
  ConvNeXt-B~\citep{convnext}& 146 & 964 & 52.7 & 45.6\\
  \rowcolor[rgb]{0.95,0.95,0.95} \textbf{PeLK-B} & 147 & 1028 &\textbf{52.9}  &  \textbf{45.9}\\
   \rowcolor[rgb]{0.95,0.95,0.95} \textbf{PeLK-B-101} & 147 & 1127 &\textbf{53.1}  &  \textbf{46.1}\\
		\bottomrule
	\end{tabular}
	}
\vspace{-3mm} 
\end{table}

\subsection{ImageNet Classification}
The ImageNet-1K~\cite{imagenet} dataset consists of 1000 object classes with 1.28M training images and 50,000 validation images. We extend the aforementioned training schedule in Section~\ref{sec3} to 300 epochs for a fair comparison. we conduct experiments for PeLK-T/S/B with input resolution $224\times224$. For PeLK-B and PeLK-B-101, we further experiment with input resolution of $384\times384$. More details of the training configurations can be found in Appendix \textcolor{red}{A}.

We compare PeLK with other state-of-the-art architectures under similar model size and FLOPs. As shown in Table~\ref{image classification}, our model outperforms powerful modern CNNs and transformers like ConvNeXt~\cite{convnext} and Swin~\cite{swin} by large margins. Notably, further scaling up the kernel size to extremely large (e.g., PeLK-101) can achieve consistent improvements. 
It is important to note that very large dense kernels are not intended for ImageNet classification, but our PeLK still exhibits a promising performance.

\begin{table}[t]
	\centering
	 \caption { Image classification accuracy (\%) comparison on ImageNet-1K. We report the top-1 accuracy. Although very large dense kernels are not intended for ImageNet classification, our PeLK still exhibits a promising performance. }
 \label{image classification}
	\scalebox{0.92}{
	\begin{tabular}[c]{lcccc}
		\toprule
            {\multirow{2}{*}{Method}} & Input & Params & FLOPs & Top-1\\  
            {} & size  & (M) & (G) & acc\\
		\midrule
  Swin-T~\citep{swin}& $224^2$ & 28 & 4.5 & 81.3\\
   T2T-Vi$\text{T}_{t}$-14~\citep{t2t}& $224^2$ & 22 & 6.1 & 81.7\\
   PerViT-S~\citep{pervit} & $224^2$ & 21 &4.4&  82.1\\
  ConvNeXt-T~\citep{convnext}& $224^2$ & 29 & 4.5 & 82.1\\
   
\rowcolor[rgb]{0.95,0.95,0.95}\textbf{PeLK-T} & $224^2$  & 29 & 5.6 & \textbf{82.6}\\
		\midrule
  PVT-Large~\citep{pvt}& $224^2$ & 61 & 9.8 & 81.7\\
   T2T-Vi$\text{T}_{t}$-19~\citep{t2t}& $224^2$ & 39 & 9.8 & 82.4\\
   PerViT-M~\citep{pervit}& $224^2$ & 44 &9.0& 82.9\\
  Swin-S~\citep{swin}& $224^2$ & 50 & 8.7 & 83.0\\

  ConvNeXt-S~\citep{convnext}& $224^2$ & 50 & 8.7 & 83.1\\
  \rowcolor[rgb]{0.95,0.95,0.95}
\textbf{PeLK-S} & $224^2$ & 50 & 10.7 &  \textbf{83.9}\\
		\midrule
  DeiT-B/16~\citep{deit}& $224^2$ & 87 & 17.6 & 81.8\\
    RepLKNet-31B~\citep{replk}& $224^2$ & 79 & 15.3 & 83.5\\
  Swin-B~\citep{swin}& $224^2$ & 88 & 15.4 & 83.5\\
  ConvNeXt-B~\citep{convnext}& $224^2$ & 89 & 15.4 & 83.8\\
  SLaK-B~\citep{slak}& $224^2$ & 95 & 17.1 & 84.0\\
  \rowcolor[rgb]{0.95,0.95,0.95}
  \textbf{PeLK-B} & $224^2$ & 89 & 18.3 &  \textbf{84.2}\\
  \midrule
  ViT-B/16~\citep{vit}& $384^2$ & 87 & 55.5 & 77.9\\
  DeiT-B/16~\citep{deit}& $384^2$ & 87 & 55.4 & 83.1\\
  Swin-B~\citep{swin}& $384^2$ & 88 & 47.1 & 84.5\\
  RepLKNet-31B~\citep{replk}& $384^2$ & 79 & 45.1 & 84.8\\
  ConvNeXt-B~\citep{convnext}& $384^2$ & 89 & 45.0 & 85.1\\
  SLaK-B~\citep{slak}& $384^2$ & 95 & 50.3 & 85.5\\
  \rowcolor[rgb]{0.95,0.95,0.95}
  \textbf{PeLK-B} & $384^2$ & 89 & 54.0 &  \textbf{85.6}\\
  \rowcolor[rgb]{0.95,0.95,0.95}
  \textbf{PeLK-B-101} & $384^2$ & 90 & 68.3 &  \textbf{85.8}\\
		\bottomrule
	\end{tabular}
	}
\vspace{-3.9mm} 
\end{table}

\subsection{Ablation Studies} \label{ablation}
\textbf{Ablation on the sharing grid.} We dive into what kind of sharing and granularity benefits most. For ease of understanding, we firstly give two instances to clearly indicate the sharing grid. For example, in Fig.~\ref{fig:left}, we parameterize a $5\times5$ convolution using a $3\times3$ convolution, where the corresponding sharing grid is $[2, 1, 2]$. Each number represents the grid size parameterized by a single parameter. For Fig.~\ref{fig:right}, we parameterize $31\times31$ convolution with a $11\times11$ convolution, the corresponding gird is $[7, 4, 2, 1, 1, 1, 1, 1, 2, 4, 7]$. Since the grid is symmetric at the center 1 (which is the central point in the kernel), we denote only half grid in Table~\ref{grid ablation} for simplicity. 

We conduct experiments with the same 120-epoch schedule on ImageNet as in Section~\ref{sec3}. We use PeLK-T without dynamic sparsity to give a sheer ablation on the sharing grid. For the baseline, we make the sharing grid to be all one (i.e., [1, 1, ..., 1]), in this way, it is equal to a $33\times33$ dense convolution as taken in RepLKNet. Results in Table~\ref{grid ablation} demonstrate that: \textbf{1)} the central fine granularity is of vital importance, while the peripheral regions can withstand wide range of sharing. \# 2, 3 show that keeping the central $5\times5$ region unshared is the key to keep performance; \# 3, 4, 5 exhibit that sharing in peripheral regions will not backfire performance evidently. We term this characteristic as focus-and-blur mechanism; \textbf{2)} an exponentially-increasing grid works best. Comparing \# 4 with \# 5, exponential gird not only reduces the parameters needed but also boosts the accuracy. From the above analysis, it can be seen that our design enjoys both the least amount of parameters and the highest performance.

\begin{table}[t]
	\centering
  \caption {Ablation study on sharing grid. No kernel-wise positional embedding is used.}
 \label{grid ablation}
 \small
	\begin{tabular}[c]{l|c|c|c}
		\toprule
		\# &{Sharing Grid} & {Param} &{Top-1 Acc}  \\
		\midrule
   1&$[1,1,...,1,1]$& 1.00$\times$ &81.4\\
    \midrule
   2&$[2,2,2,2,2,2,2,2,1]$ & 0.27$\times$ &81.0\\
   3&$[2,2,2,2,2,2,2,1,1,1]$& 0.33$\times$ &81.4\\		
   4&$[4,4,4,2,1,1,1]$& 0.16$\times$ &81.3\\
\rowcolor[rgb]{0.95,0.95,0.95}   5&$[8,4,2,1,1,1 ]$& 0.11$\times$ &81.4\\
   6&$[1,1,2,4,8,1 ]$& 0.11$\times$ &80.5\\
 
		\bottomrule
	\end{tabular}
\vspace{-3mm} 
\end{table}

\textbf{Ablation on the central fine-grained area ratio.} 
Table~\ref{central ablation} ablates the effect of varying central fine-grained kernel size (i.e., the focus region). We also report the proportion of the central region to the total kernel size. The results show that the central region only takes about 1\% proportion to maintain the model's high performance. However, the central region can not be too small, which will lead to severe performance degradation. Further increasing the central region does not bring additional benefits, but it brings additional parameters. In our main experiments, we keep the central $5\times5$ region of PeLK as fine-grained, and for PeLK-101, we enlarge the central region to $7\times7$ to maintain a similar central ratio.

\begin{table}[]
	\begin{center}
   \caption{Ablation on the central fine-grained kernel size. Kernel-wise positional
embedding is used.} 
  \label{central ablation}
		\small
		\begin{tabular}{c|c|c|c}
			\toprule
			Sharing Grid			&	Central	Kernel&	Ratio	&	Top-1 Acc \\
                \midrule

                $[11,8,4,2,1]$      &   $1\times1$    &   0.04\%        & 80.8 \\
                
               $[10,8,4,2,1,1]$      &   $3\times3$     &   0.35\%        & 81.1 \\

\rowcolor[rgb]{0.95,0.95,0.95}				$[9,8,4,2,1,1,1]$       &   $5\times5$     &   0.96\%         & 81.6 \\
		$[8,8,4,2,1,1,1,1]$     &   $7\times7$    &   1.88\%         & 81.6 \\
			\bottomrule
		\end{tabular}	
  \end{center}
\vspace{-5mm}
\end{table}

\begin{table}[b]
	\begin{center}
   \caption{Ablation on the kernel size configuration. Kernel-wise positional
embedding is used.} 
  \label{configuration ablation}
		\small
		\begin{tabular}{l|c|c|c}
			\toprule
			Model			&	Input Size &	Kernel Size	&	Top-1 Acc \\
                \midrule

                PeLK      &   $224\times224$    &   51-49-47-13        & 81.6 \\
                
               PeLK-101      &   $224\times224$    &   101-69-67-13        & 81.6 \\
               PeLK-151      &   $224\times224$    &   151-89-87-13        & 81.2 \\
\midrule
PeLK      &   $384\times384$     &   51-49-47-13         & 82.7 \\
		PeLK-101    &   $384\times384$  &   101-69-67-13        & 83.0 \\
  PeLK-151    &  $384\times384$   &   151-89-87-13        & 82.8 \\
			\bottomrule
		\end{tabular}	
  \end{center}
\vspace{-0.15in}
\end{table}

\textbf{Ablation on the kernel configuration.} Table~\ref{configuration ablation} ablates the configuration of kernel size in a 120 epoch schedule as in Section~\ref{sec3}. For the input resolution of $224^2$, enlarging kernel size to $101\times101$ will not bring additional benefits; while for input resolution of $384^2$, PeLK-101 obtains a clear advantage over PeLK. Increasing kernel size to $152\times152$ leads to performance degradation, especially for input resolution of $224^2$. These phenomena are reasonable considering the input resolution. For a typical convnet like ConvNeXt or our PeLK, the stem layer will result in a $4\times$ downsampling of the input images. So for input $224^2$, a $51\times51$ kernel is roughly able to cover the global feature map after stem. And for input $384^2$, a $101\times101$ kernel is equal to a global convolution, thus continuing scaling up kernel can not bring more global perception but only wasted parameters. This essentially suggests that kernel configuration should be tightly related to the input size. Currently, for the most commonly used $224^2$ and $384^2$ training, PeLK and PeLK-101 are the suitable options respectively. Moreover, with the development of hardware devices and computing power in the future, our approach will hopefully shine further when it is affordable to pretrain at higher resolutions.

\section{Analysis}
\begin{figure}[t]
\centering
\includegraphics[width=1.0\linewidth]{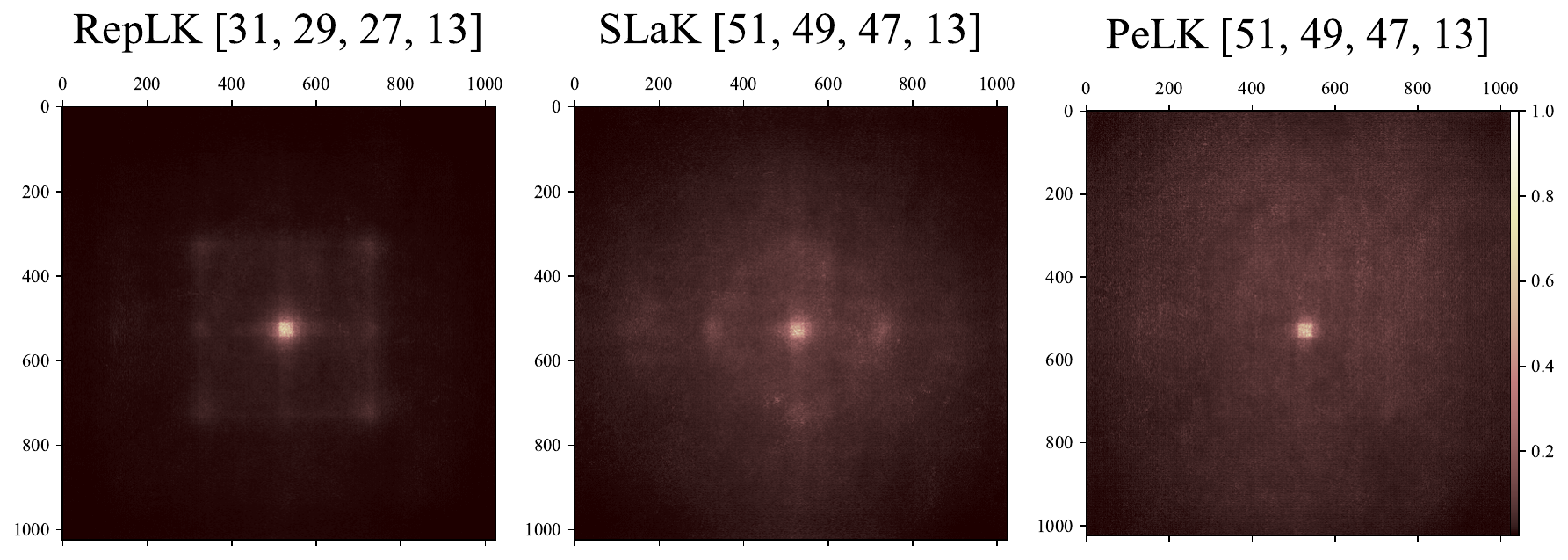} 
\caption{\textbf{Effective receptive field (ERF) comparison.} Our PeLK has larger ERFs than SLaK and RepLK, spreading a wider area.}
 \label{erf}
\vspace{-3pt} 
\end{figure}
\subsection{Visualization of ERFs.} 
Previous large kernel convnets like RepLKNet and SLaK attribute their performance gains to their large Effective Receptive Fields (ERFs). Facilitated by peripheral convolution, PeLK has a much larger perception range. Therefore, we argue that PeLK's strong performance comes from larger ERFs.
To verify, we depict the ERFs following RepLKNet and SLaK, we sample and resize 50 images from the validation set to $1024\times1024$,
and measure the contribution of the pixel on input images to the central point of the feature map
generated in the last layer. The contribution scores are further accumulated and projected to a
$1024\times1024$ matrix, as visualized in Fig~\ref{erf}. Our PeLK spreads high-contribution pixels in a much larger ERF, validating our hypothesis and further exhibiting our effectivess.

\begin{table}[b]
	\centering
  \caption {Inference throughput comparison on ImageNet-1K. The results are in FP32 precision. We use an A100 GPU with PyTorch 1.10.0 + cuDNN 8.2.0 to conduct this experiment.}
 \label{inference throughput comparison}
 \scalebox{0.92}{
	\begin{tabular}[c]{l|c|c|c}
		\toprule
		Models & Input& Kernel Size &Throughput \\
		\midrule
   SLaK-T~\cite{slak}& $224^2$&51-49-47-13 &754\\
   \rowcolor[rgb]{0.95,0.95,0.95}
   PeLK-T&$224^2$&51-49-47-13 &\textbf{1138}\\		
   \rowcolor[rgb]{0.95,0.95,0.95}
   PeLK-101-T& $224^2$&101-69-67-13 &\textbf{1077}\\
		\bottomrule
	\end{tabular}}
\vspace{-3mm} 
\end{table}

\begin{figure}[t]
\centering
\includegraphics[width=1.0\linewidth]{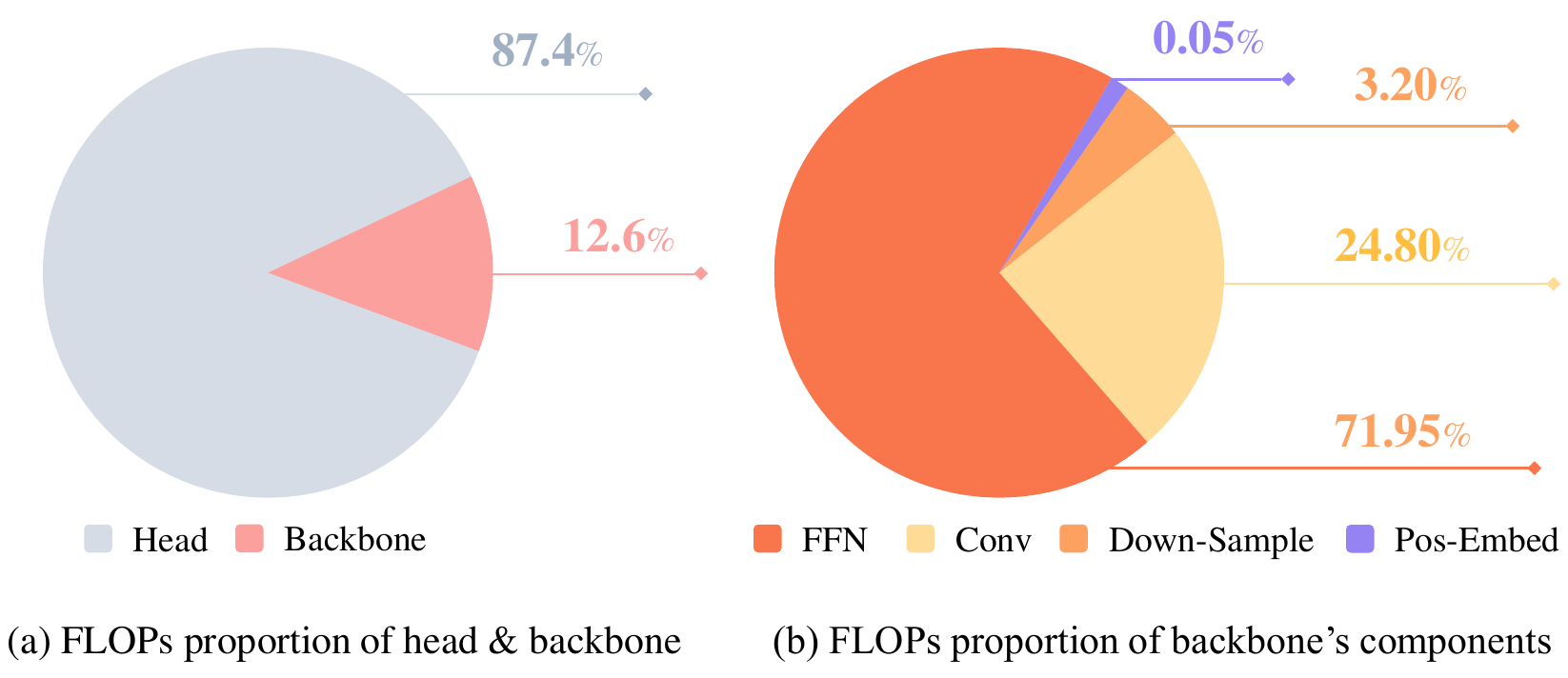} 
\caption{\textbf{Analysis of FLOPs.} (a) FLOPs proportion of head \& backbone. (b) FLOPs proportion of backbone's components. The head is UperNet and the backbone is PeLK-T respectively. FLOPs are based on input sizes of (2048, 512).}
 \label{FLOPs}
\vspace{-5mm} 
\end{figure}
\subsection{Analysis of FLOPs}
We provide a detailed breakdown of the FLOPs for the PeLK-T architecture utilized in semantic segmentation in Fig.\ref{FLOPs}. As shown in Fig.\textcolor[rgb]{1,0,0}{5(a)}, we depict the FLOPs distribution between the head (i.e., UperNet~\cite{upernet}) and backbone (i.e., PeLK-T) of the model. In Fig.\textcolor[rgb]{1,0,0}{5(b)}, we give a comprehensive analysis of the FLOPs contributions from different components of the backbone (i.e., PeLK-T), including FFNs, large-kernel convolutions, down-sampling layers, and kernel-wise positional embedding. There are two noteworthy points. Firstly, large kernel convolutions account for approximately 25\% of the overall FLOPs of the backbone, thus further scaling up the kernel size does not significantly increase the overall FLOPs. Secondly, the extra FLOPs introduced by positional embedding are minimal, accounting for only 0.05\% of the backbone's FLOPs. So, kernel-wise positional embed is both cheap and elegant. 

\subsection{Inference Throughput Measurement}

We compare inference throughput measurement in Table~\ref{inference throughput comparison}. The results are obtained on an A100 GPU with input resolution of $224\times224$. We use PyTorch 1.10.0 + cuDNN 8.2.0 and FP32 precision. Although SLaK uses stripe convolution to speed up the computation of very large kernel, we still hold a clear speed advantage (i.e., $1.5\times$ speedup). This advantage is particularly remarkable considering that PeLK outperforms SLaK on ADE20K, COCO and ImageNet. More importantly, scaling up kernel to 101 only brings minor speed overhead, further exhibiting our design's merits in scaling properties.

\subsection{Kernel Scaling Efficiency.} 
\begin{figure}[t]
\centering
\includegraphics[width=0.96\linewidth]{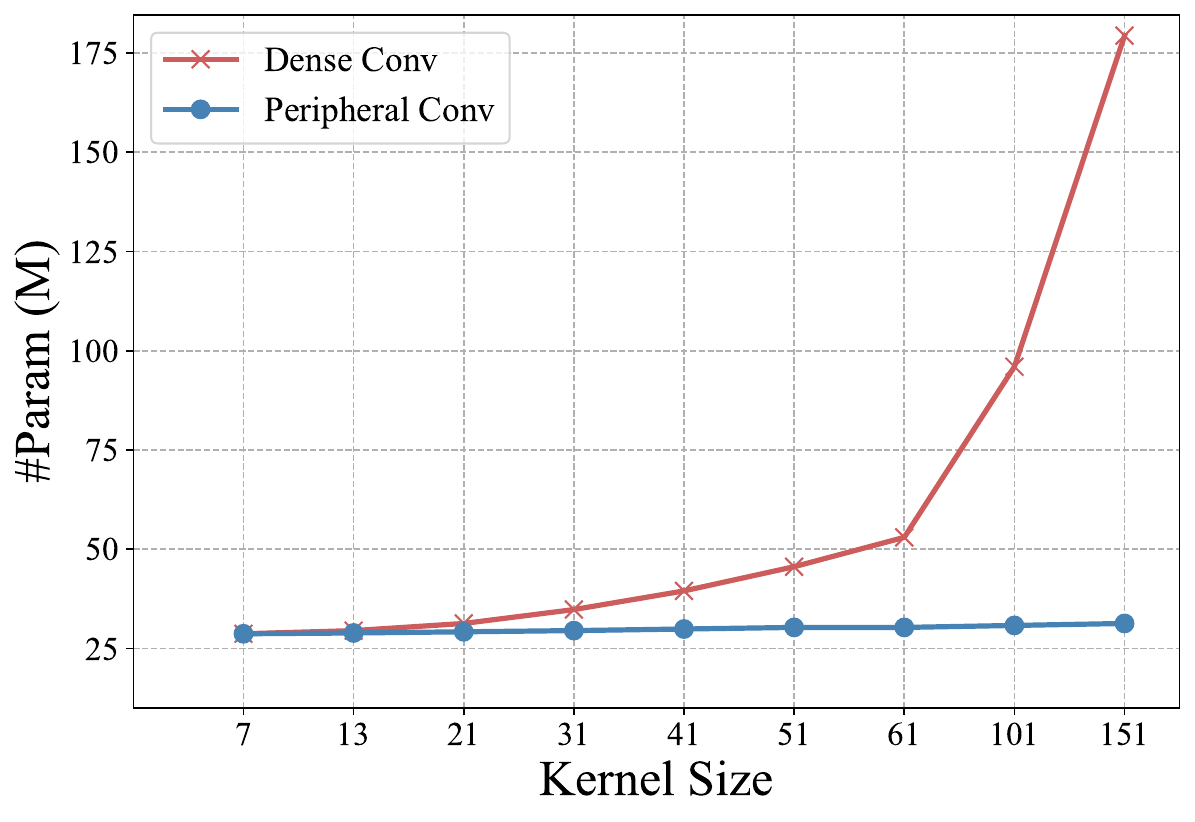} 
\caption{\textbf{Scaling efficiency comparison.} We compare the model size with a set of kernel sizes from 7 to 151. Our peripheral convolution has a clear advantage, bringing minor parameter overhead.}
 \label{scaling efficiency}
\vspace{-5mm} 
\end{figure}

Our peripheral convolution reduces
the parameter complexity of dense convolutions from $O(K^2)$ to $O(\log{K})$, which enables us to scale up kernel size with a remarkably minor model size overhead. To demonstrate this, we simply replace all the kernels in stages of ConvNeXt-T with a set of kernel sizes from 7 to 151 and report the required number of parameters. As shown in Fig~\ref{scaling efficiency}, our approach exhibits a remarkable scaling advantage, and we can see a clear gap when the kernel size is larger than 50. Using dense convolution results in a rapidly growing model size, which is unacceptable in practice. In contrast, our peripheral convolution incurs only a minor model size overhead, making it possible to design extremely large kernel convnets.

\section{Conclusion}
This paper explores the design of extremely large kernel convolutional neural networks. We propose a new form of convolution termed peripheral convolution, which can reduce the parameter complexity of dense convolution from $O(K^2)$ to $O(\log{K})$ while keeping dense convolution's merits. Built upon the proposed peripheral convolution, we design extremely large dense kernel CNNs and achieve notable improvements across a variety of vision tasks. Our strong results suggest biologically inspired mechanisms can make a promising tool to boost modern network design.

\section*{Acknowledgments}
This work is supported in part by the National Key R\&D Program of China (Grant No.2022ZD0116403), the National Natural Science Foundation of China (Grant No. 61721004), and the Strategic Priority Research
Program of Chinese Academy of Sciences (Grant No.
XDA27000000).

We thank Yurong Zhang for the help in the  depiction of Fig.1(b) and Bo Zhang for technical support.
\newpage

{
    \small
    \bibliographystyle{ieeenat_fullname}
    \bibliography{main}
}

\clearpage
\setcounter{section}{0}
\renewcommand\thesection{\Alph{section}}

\section{ImageNet Training Configuration}
For training PeLK, we use 8 GPUs and a total batch size of 4096 to train for 300 epochs. The optimizer is AdamW~\cite{adamw} with momentum of 0.9 and weight decay of 0.05. The learning rate setting includes an initial value of $4\times10^{-3}$, cosine annealing and 20-epoch warm-up. For the data augmentation and regularization, we use RandAugment~\cite{randaug} (“rand-m9-mstd0.5-inc1” as implemented by timm~\cite{timm}), label smoothing coefficient of 0.1, mixup~\cite{mixup} with $\alpha = 0.8$, CutMix~\cite{cutmix} with $\alpha = 1.0$, Rand Erasing~\cite{randerase} with probability of 25\% and Stochastic Depth with a drop-path rate of 10\%/40\%/50\% for PeLK-T/S/B respectively.

\section{ERF Quantitation Comparison}
Following RepLKNet~\cite{replk} and SLaK~\cite{slak}, we report the high-contribution area ratio $r$ to give a quantitation analysis of ERF comparison in Table \textcolor{red}{9}. Here, $r$ denotes the proportion of the smallest rectangle to the overall input area that can encompass the contribution scores above a specified threshold $t$. For instance, given an area of $\rm R\times R$ at the center can cover $\rm t = 20\%$ contribution scores of a $1024\times1024$ input, the corresponding area ratio of t = 20\% is $r = (\rm R/1024)^2$. Larger $r$ indicates a smoother distribution of high-contribution pixels. Compared with previous CNN paradigms, our PeLK naturally takes a larger range of pixels into accout to make decisions, which continues to demonstrate the intuitive effect of the extremely large kernel on enlarging the receptive field.
\begin{table}[h]
	\centering
	\label{quantitation of erfs}
	\begin{tabular}[c]{l|c|ccc}
		\toprule
		Models & Kernel Size &t=20\%& t=30\%& t=50\% \\
		\midrule
  ResNet& 3-3-3-3 &1.1\% & 1.8\% &3.9\%\\
  ConvNeXt& 7-7-7-7 &2.0\% & 3.6\% &7.7\%\\
   RepLKNet& 31-29-27-13 &4.0\% & 9.1\% &19.1\%\\
   SLaK& 51-49-47-13 &6.9\% & 11.5\% &23.4\%\\
   PeLK&51-49-47-13 &\textbf{7.5\%} & \textbf{12.8\%} &\textbf{25.9\%}\\		
   PeLK-101& 101-69-67-13 &\textbf{8.1\%} & \textbf{13.7\%} &\textbf{26.5\%}\\

		\bottomrule
	\end{tabular}
 \caption {\textbf{Quantitative comparison of ERF.} We use ResNet-152 and tiny size model for the other methods. larger values indicate larger ERFs and smoother distribution of high-contribution pixels.}
\end{table}

\section{Ablation on Re-parameterization}
According to RepLKNet~\cite{replk}, directly optimizing large kernel convnets can be difficult and leads to performance degradation. Therefore, existing large kernel paradigms re-parameterize a small kernel (e.g., $5\times5$) to alleviate this issue. In this part, we remove the re-parameterization trick to see how degraded the models are. We train tiny model for 120 epochs on ImageNet as the same in Section \textcolor{red}{3}. As shown in Table \textcolor{red}{10}, our peripheral convolution still sustain a good performance after removing the small convolution, while the dense convolution suffers a significant degradation. This phenomenon implies that our peripheral convolution can alleviate the optimization difficulty of large dense convolution by reducing the number of parameters required.

\begin{table}[h]
	\centering
	\label{easier optimization}
	\begin{tabular}[c]{l|c|c|c|c}
		\toprule
		Models & Conv Form& w/ Rep & w/o Rep & $\Delta$ \\
		\midrule
   RepLK& dense &81.6&80.2& -1.4\\		
   PeLK& peripheral&81.6 &80.9&\textbf{-0.7}\\

		\bottomrule
	\end{tabular}
 \caption {\textbf{Ablation on Re-parameterization.} We compare the degradation of the model after removing the rep technique.}
\end{table}

\begin{figure}[t]
\centering
\includegraphics[width=1.0\linewidth]{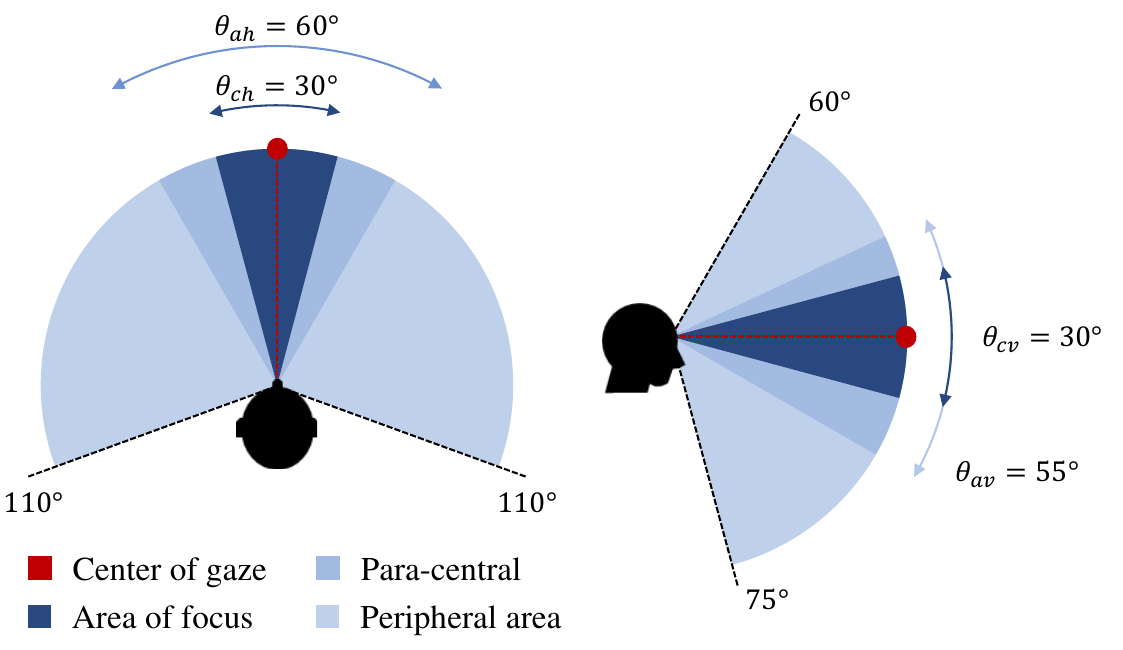} 
\caption{\textbf{Illustration of peripheral vision.} Human vision possesses distinct clarity within a confined focus area, contrasted by merely vague perception in the extensive peripheral area. Note that all numbers are approximate values.}
 \label{peripheral vision}
\vspace{-15pt} 
\end{figure}

\section{Peripheral Vision} 
The human visual field showcases a phenomenon known as ”central focus and peripheral blur”~\cite{min2022peripheral,pramod2022human}. According to vision science literature\cite{marieb2007human, strasburger2011peripheral}, the human visual field can be modeled as fan-shaped figures as shown in Fig~\ref{peripheral vision}. It consists of three segments from the center to the periphery: central area (${\theta_{ch}}$ and ${\theta_{cv}}$), para-central area (${\theta_{ah}}$ and ${\theta_{av}}$) and peripheral area (${\theta_{ph}}$ and ${\theta_{pv}}$). The central area is the primary part used for clear perception. Therefore, the proportion of the focused area in the human visual system can be calculated as:
\begin{equation}
	P_{human} = \frac{\pi \theta_{ch} \cdot \theta_{cv}}{\pi \theta_{ph} \cdot \theta_{pv}} =2.72\%
\end{equation}
Similarly, our peripheral convolution keeps fine-grained parameters in the center, the central proportion for PeLK is:
\begin{equation}
	P_{pelk} = \frac{5\times5}{51\times51} =0.96\%
\end{equation}
Although the values for PeLK and human are not strictly equivalent, they both are very small ratios ($<5\%$), indicating that an efficient visual mechanism only requires a very small proportion of fine-grained perception.

\end{document}